\documentclass[11pt]{article}
\usepackage[final]{acl}

\usepackage{times}
\usepackage{latexsym}

\usepackage{amsmath}
\usepackage{booktabs}
\usepackage{multirow}
\usepackage{tcolorbox}
\tcbuselibrary{breakable, skins}
\usepackage{listings}
\usepackage[T1]{fontenc}

\usepackage[utf8]{inputenc}

\usepackage{microtype}

\usepackage{inconsolata}

\usepackage{graphicx}

\title{ProductResearch: Training E-Commerce Deep Research Agents via Multi-Agent Synthetic Trajectory Distillation}

\author{
 \textbf{Jiangyuan Wang\textsuperscript{1}}$^{*}$,
 \textbf{Kejun Xiao\textsuperscript{1}}$^{*\dagger}$, 
\\
 \textbf{Huaipeng Zhao\textsuperscript{1}},
 \textbf{Tao Luo\textsuperscript{1}},
 \textbf{Xiaoyi Zeng\textsuperscript{1}}
\\
 \textsuperscript{1}Alibaba International Digital Commercial Group,
\\
 \small{
   \textbf{Correspondence:} \href{mailto:email@domain}{{wangjiangyuan.wjy,xiaokejunkejun.xia}@alibaba-inc.com}
 }
}

\begin{document}
\maketitle
\renewcommand{\thefootnote}{}
\footnotetext{$^{*}$Equal contribution.}
\footnotetext{$^{\dagger}$Corresponding author.}
\renewcommand{\thefootnote}{\arabic{footnote}}
\begin{abstract}
Large Language Model (LLM)-based agents show promise for e-commerce conversational shopping, yet existing implementations lack the interaction depth and contextual breadth required for complex product research. Meanwhile, the Deep Research paradigm, despite advancing information synthesis in web search, suffers from domain gaps when transferred to e-commerce. We propose \textit{ProductResearch}, a multi-agent framework that synthesizes high-fidelity, long-horizon tool-use trajectories for training robust e-commerce shopping agents. The framework employs a User Agent to infer nuanced shopping intents from behavioral histories, and a Supervisor Agent that orchestrates iterative collaboration with a Research Agent to generate synthetic trajectories culminating in comprehensive, insightful product research reports. These trajectories are rigorously filtered and distilled through a reflective internalization process that consolidates multi-agent supervisory interactions into coherent single-role training examples, enabling effective fine-tuning of LLM agents for complex shopping inquiries.  Extensive experiments show that a compact MoE model fine-tuned on our synthetic data achieves substantial improvements over its base model in response comprehensiveness, research depth, and user-perceived utility, approaching the performance of frontier proprietary deep research systems and establishing multi-agent synthetic trajectory training as an effective and scalable paradigm for enhancing LLM-based shopping assistance.
\end{abstract}

\section{Introduction}

The rapid evolution of LLM-based agents \cite{yaoReActSynergizingReasoning2023a} has catalyzed a paradigm shift in e-commerce, where conversational shopping agents increasingly mediate consumer decision-making \cite{yaoWebShopScalableRealWorld2022, wangShopSimulatorEvaluatingExploring2026}. Modern users frequently confront complex, information-intensive purchasing decisions that span both consumer and business perspectives—a customer selecting a professional camera system for specific environmental conditions, or a merchant analyzing current market trends in baby products to identify what's gaining traction. These scenarios demand not merely item retrieval or simple recommendation, but sustained, multi-source research culminating in comprehensive, evidence-grounded analysis. The ReAct-style approach that interleave reasoning and action to building such agents has proven effective for straightforward shopping tasks \cite{wangShoppingBenchRealWorldIntentGrounded2025}, yet its potential for complex product research remains largely underexplored. Existing frameworks \cite{yaoWebShopScalableRealWorld2022, fangMultiAgentConversationalRecommender2024}, while advancing task completion and conversational recommendation, remain oriented toward binary success metrics rather than the informational depth and evidentiary rigor that complex purchasing decisions demand.

\begin{figure*}[t]
  \centering
  \includegraphics[width=\textwidth]{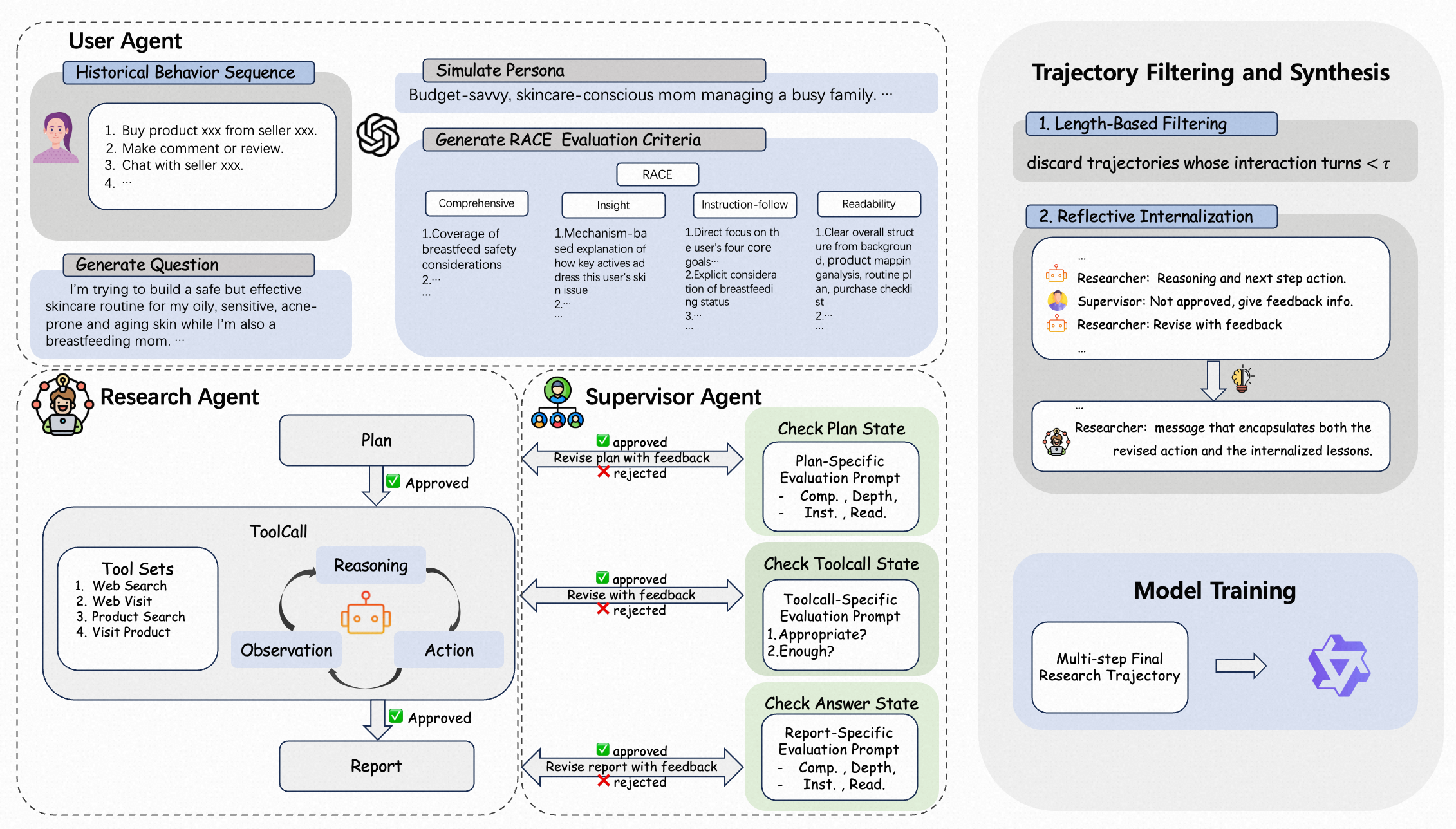}
  \caption{Overview of the ProductResearch framework. The User Agent infers a persona, research query, and RACE evaluation rubric from behavioral histories. The Research Agent executes Plan→Toolcall→Report reasoning under step-level supervision from the Supervisor Agent, which provides state-specific verification and corrective feedback. Approved trajectories are filtered by length and distilled via reflective internalization into single-role sequences for supervised fine-tuning.}
  \label{fig:product_research_architecture}
\end{figure*}

Meanwhile, the emerging "Deep Research" paradigm has achieved remarkable success in open-domain information synthesis, enabling agents to conduct extended, multi-step investigations that rival human-expert analysis in web search scenarios \cite{teamTongyiDeepResearchTechnical2025, shaoDRTuluReinforcement2025, liWebWeaverStructuringWebScale2025}. These systems orchestrate iterative planning, evidence acquisition, and report generation over long horizons, producing richly detailed and well-structured outputs\cite{qiaoWebResearcherUnleashingUnbounded2025,Qwen-DeepResearch, GeminiDeepResearch}. However, as noted in the Tongyi DeepResearch technical report \cite{teamTongyiDeepResearchTechnical2025}, such models are primarily optimized for web search tool use and lack robustness for broader agentic tool use scenarios. Our investigation also reveals that this paradigm does not transfer effectively to e-commerce. When applied to complex shopping inquiries, Deep Research agents encounter significant domain generalization challenges, as e-commerce demands seamless orchestration of open-web knowledge gathering with structured product catalog querying, grounding of claims in verified product attributes, and synthesis of heterogeneous evidence sources (expert reviews, user feedback, technical specifications). This finding motivates our work on furthering the potential of ReAct-style agents in this domain, equipping them with the analytical depth, contextual breadth, and evidentiary rigor needed for complex pre-purchase research. 

To bridge this gap, we propose ProductResearch, a multi-agent framework that synthesizes high-fidelity, long-horizon tool-use trajectories to train robust e-commerce deep research agents. Our framework comprises three specialized agents working in concert. A User Agent, grounded in real user behavioral histories, infers nuanced shopping intents and generates both complex research queries and query-adaptive evaluation rubrics. A Research Agent, equipped with a dual-environment toolset spanning both the open web and a large-scale product catalog, executes extended research trajectories through iterative reasoning and tool interaction. Critically, a Supervisor Agent, governed by a three-stage state machine, provides step-level quality supervision across the plan–toolcall–report lifecycle, detecting and correcting hallucinations, logic drift, and insufficient evidence coverage through targeted feedback. The resulting trajectories undergo a reflective internalization process that distills multi-turn supervisory interactions into coherent single-role training examples, preserving the corrective learning signals while enabling standard supervised fine-tuning.

Extensive experiments validate the effectiveness of our approach. After fine-tuning on ProductResearch-generated trajectories, a compact mixture-of-experts (MoE) model (Qwen3-30B-A3B) improves its overall RACE score from 31.78 to 45.40 with consistent improvements across all evaluation dimensions. The fine-tuned model also achieves an effective product coverage of 12.45, more than tripling its base model's score of 3.58, demonstrating enhanced breadth in product investigation alongside higher report quality. These results confirm that high-quality synthetic trajectories generated by our multi-agent framework can effectively internalize complex research behaviors into lightweight models, establishing a scalable paradigm for building capable e-commerce deep research agents. Our principal contributions are summarized as follows:
\begin{itemize}
  \setlength{\itemsep}{0pt}      
  \setlength{\parsep}{0pt}       
  \setlength{\parskip}{0pt}      
    \item We introduce a novel product research dataset with complex queries, evaluation rubrics, and agent trajectories, serving as both a training corpus and a benchmark for evaluating product research report capabilities. 
    
    \item We propose ProductResearch, a novel multi-agent framework for scalable synthesis of high-fidelity e-commerce research trajectories.

    \item We demonstrate through extensive experiments that LLM agents fine-tuned on ProductResearch-generated synthetic trajectories achieve significant improvements across all evaluation dimensions.
\end{itemize}

\section{Method}
In this section, we detail the \textit{ProductResearch} framework, a multi-agent system designed to synthesize high-fidelity, long-horizon tool-use trajectories for e-commerce shopping research. The framework consists of three specialized agents: the User Agent, the Research Agent, and the Supervisor Agent. The interaction between these agents is governed by a state-machine-guided feedback loop to ensure the logical consistency and domain-specific accuracy of the generated data.

\subsection{Overview of the Multi-Agent Framework}
The core objective of our framework is to emulate the ``Deep Research'' paradigm within the e-commerce domain. Unlike standard ReAct-style trajectories, our method generates extended interaction logs where an agent must synthesize external web knowledge with internal product database queries. As illustrated in Figure~\ref{fig:product_research_architecture}, the system workflow is formalized as an iterative optimization process across three distinct phases: (1) User Profiling and Query Formulation, (2) State-Aware Supervised Iterative Research Execution and (3) Trajectory Refinement and Distillation. The complete set of prompts used by all agents throughout the framework is provided in Appendix~\ref{app:prompts}.

\subsection{Phase 1: User Profiling and Query Formulation}
To ensure the synthetic data reflects real-world diversity, we initialize the process by grounding the \textbf{User Agent} in actual user behavior.

\textbf{Persona and Need Extraction.} Given a user's long-term historical behavioral sequence $S = \{b_1, b_2, \dots, b_n\}$ (e.g., purchases, reviews, conversations with platform or seller), the User Agent generates a multidimensional user profile $\mathbf{P}$ and a specific research query $\mathbf{Q}$. The query is designed to be a complex information-seeking task that cannot be answered by simple retrieval.

\textbf{Dynamic Evaluation Criteria Generation.} Crucially, the User Agent also generates a set of dynamic evaluation criteria along four dimensions: comprehensiveness, depth, instruction-following, and readability. For each query $\mathbf{Q}$, the User Agent assigns dimension-level weights $\mathbf{W} = \{W_1, \dots, W_4\}$ and, within each dimension $k$, criterion-level weights $\mathbf{w}_k$, providing a customized rubric for the Supervisor Agent to utilize during trajectory generation.

\subsection{Phase 2: State-Aware Supervised Iterative Research Execution}

The \textbf{Research Agent}, instantiated as an LLM operating in a ReAct-style loop, is tasked with resolving $\mathbf{Q}$ through iterative reasoning and tool interaction. While the agent retains the flexibility to dynamically interleave thought and action steps, its overall behavior is guided by a high-level cognitive schema of \textbf{Plan $\rightarrow$ Toolcall $\rightarrow$ Report}, ensuring that each research session progresses from strategic planning through evidence gathering to final report synthesis. The agent is equipped with a specialized e-commerce toolset $\mathcal{T}$ for open-web information gathering and internal product catalog querying (detailed specifications in Appendix~\ref{app:tools}).

The Research Agent must conduct thorough broad knowledge acquisition and in-depth product comparisons throughout its reasoning chain. The primary bottleneck in synthetic data quality is the hallucination or logic-drift common in long-horizon LLM outputs. We address this by introducing a Supervisor Agent governed by a three-stage state machine.

\subsubsection{State-Specific Verification}
The \textbf{Supervisor Agent} monitors the Research Agent's output at every step. It operates in three distinct states:
\begin{description}
  \item[\textbf{Check Plan}] Evaluates logical soundness and coverage completeness of the proposed research strategy against the User Agent's profile and query requirements.
  \item[\textbf{Check Toolcall}] Validates tool parameter correctness, relevance of retrieved information, and prevention of repetitive execution loops.
  \item[\textbf{Check Report}] Verifies adherence to the evaluation criteria (comprehensiveness, depth, instruction-following, and readability) and delivery of an evidence-based product investigation report.
\end{description}

\subsubsection{Iterative Feedback Loop}
The Research Agent operates across three states---\textit{Plan}, \textit{Toolcall}, and \textit{Report}---each of which may comprise multiple steps. At each step $j$ within state $S_i$, the Research Agent produces an output $O_{i,j}$ and proposes whether to remain in the current state (e.g., issuing additional tool calls based on intermediate observations) or transition to the next state. Every step is subject to inspection by the Supervisor Agent using a state-specific prompt $\Phi_i$ that encodes the User Agent's requirements and the Research Agent's behavioral guidelines. If the Supervisor detects a sub-optimal output, it generates textual feedback $F_{i,j}$, prompting the Research Agent to revise its current step. Formally:

\begin{equation}
\small
O'_{i,j} \leftarrow
\begin{cases}
O_{i,j} & \text{if } \mathcal{S}(O_{i,j}, \Phi_i) = \text{Approve} \\[4pt]
\mathcal{R}(O_{i,j}, F_{i,j}) & \text{if } \mathcal{S}(O_{i,j}, \Phi_i) = \text{Not Approve}
\end{cases}
\end{equation}

where $\mathcal{S}$ denotes the Supervisor Agent's evaluation function, $\mathcal{R}$ denotes the revision operation, and $F_{i,j}$ denotes the textual feedback generated upon rejection. In the latter case, the Research Agent revises its action for the current step based on $F_{i,j}$ before proceeding. This step-level supervision mechanism ensures that only high-fidelity, verified trajectories are retained for the final training set.

\subsection{Phase 3: Trajectory Refinement and Distillation}
The raw trajectories produced by the multi-agent interaction loop undergo two critical post-processing stages before being used for model training.

\textbf{Length-Based Filtering.} We first discard trajectories whose total number of interaction turns falls below a predefined minimum threshold $\tau$ (see Appendix~\ref{app:training} for the specific value). This filtering step eliminates trivially resolved queries and guarantees sufficient depth of reasoning and tool engagement in the retained training data.

\textbf{Trajectory Distillation via Reflective Internalization.} A more consequential challenge arises from the structural composition of the raw trajectories. During the iterative feedback loop, when the Supervisor Agent approves a step, its approval message is removed from the trajectory, as it carries no corrective signal. However, when the Supervisor rejects a step, its feedback message $F_{i,j}$ is retained alongside the Research Agent's subsequent revision. This results in interleaved multi-role sequences of the form [\textit{assistant, supervisor, assistant, \dots}], which cannot be directly consumed by standard single-role supervised fine-tuning pipelines.

To resolve this structural mismatch, we introduce a \textit{reflective internalization} step. For each such interleaved sequence, the Research Agent is prompted to review the full trajectory---including the Supervisor's feedback and its own revised actions---and to distill the corrective insights into a single, consolidated assistant message. Concretely, the agent summarizes what was initially inadequate, how the feedback guided its revision, and what the improved reasoning or action should be, then produces a self-contained output that encapsulates both the final decision and the internalized lessons. This process is analogous to how a human expert retrospectively reviews their workflow, reflects on missteps, and consolidates the experience into refined expertise. The resulting trajectories consist exclusively of coherent single-role assistant turns, making them directly amenable to supervised fine-tuning while preserving the corrective learning signals from the Supervisor Agent's quality control.

\begin{table*}
  \centering
  \small
  \renewcommand{\arraystretch}{1.15}
  \setlength{\tabcolsep}{6pt}
  \begin{tabular}{llcccccc}
  \toprule
    \multirow{2}{*}{\textbf{Category}} &
    \multirow{2}{*}{\textbf{Models}} & 
    \multicolumn{5}{c}{\textbf{RACE}} & 
    \multirow{2}{*}{\textbf{E.Prod}} \\
    \cmidrule(lr){3-7}
    & & \textbf{Overall} & \textbf{Comp.} & \textbf{Depth} & \textbf{Inst.} & \textbf{Read.} & \\
    \midrule
    \multirow{3}{*}{\shortstack[l]{\textbf{Deep Research}}}
    & Tongyi-DeepResearch & 29.84 & 29.10 & 26.43 & 33.00 & 32.79 & 6.69 \\
    & Qwen-DeepResearch & 42.76 & 41.70 & 42.87 & 43.45 & \underline{43.15} & \underline{14.4} \\
    & Gemini-DeepResearch & \textbf{45.56} & \textbf{45.81} & \textbf{47.46} & \underline{45.38} & 42.31 & \textbf{25.2} \\
    \midrule
    \multirow{4}{*}{\shortstack[l]{\textbf{ReAct}}}
    & Gemini-3-flash	& 32.41 & 30.16 & 29.17 & 38.43	& 33.85 & 6.54 \\
    & GPT-4.1 & 36.46 & 33.88 & 41.47 & 41.10 & 37.65 & 7.98 \\
    & Qwen3-max & 36.67 & 35.40 & 33.44 & 41.28 & 38.74 & 6.06 \\
    & Qwen3-30B-A3B & 31.78 & 29.81 & 28.41 & 36.33 & 35.42 & 3.58 \\
    \midrule
    \textbf{Ours}
    & ProductResearch-SFT-128k & \underline{45.40} & \underline{45.44} & \underline{43.87} & \textbf{46.09} & \textbf{47.22} & 12.45 \\
  \bottomrule
  \end{tabular}
  \caption{\label{tab:performance}
    Main results on the e-commerce product research benchmark. 
  We report the overall RACE score and four dimension-level 
  sub-scores, as well as the average Effective Product Count.  Qwen-DeepResearch and Gemini-DeepResearch are proprietary systems operating with their native built-in tools; all other models share the same tool set $\mathcal{T}$.  The best results are in \textbf{bold} 
  and the second-best are \underline{underlined}.
  }
\end{table*}


\section{Experiments}
\subsection{Experimental Settings}
\noindent\textbf{Dataset Construction.}
We collected anonymized real-world user interaction logs—including purchase histories, reviews, and customer service dialogues—and curated 1,000 representative users to instantiate the User Agent's persona simulation. The final dataset was partitioned into training, validation, and test sets following an 8:1:1 ratio.

\noindent\textbf{Baselines.}
We compare against two categories of baselines. (1) \textit{Deep Research Agents}: Tongyi-DeepResearch, an open-source model for which we deploy the same tool set $T$ as our method; and two proprietary systems, Qwen-DeepResearch and Gemini-DeepResearch, which operate with their native built-in tool capabilities. (2) \textit{ReAct Agents}: we equip three frontier LLMs—Gemini-3-flash, GPT-4.1, and Qwen3-max—with the same tool set $T$ and deploy them in a standard ReAct loop. We additionally include Qwen3-30B-A3B under this setting as the base model of our fine-tuned variant, enabling direct assessment of the synthetic training data's contribution. We fine-tuned Qwen3-30B-A3B with context-length variants ranging from 32k to 128k tokens on a 32×A100 GPU cluster using Megatron-LM. Full training details are documented in Appendix~\ref{app:training}.

\subsection{Evaluation of Product Investigation Report}

We adopt the \textbf{RACE} metric from Deep Research Bench~\cite{duDeepResearchBenchComprehensive2025} and adapt it to the e-commerce product research setting. RACE performs query-adaptive, rubric-driven pairwise comparison of report quality, scoring a target report against a reference report across multiple dimensions and aggregating via hierarchical weighted summation into a final score (detailed formulation in Appendix~\ref{app:race}).

\paragraph{Query-Specific Evaluation Rubrics.} We leverage the User Agent to dynamically generate fine-grained evaluation criteria and dimension-level weights tailored to each user query, ensuring that reports are judged against the specific information needs of the underlying shopping intent.

\paragraph{Reference Report Construction.} For each test query, we designate the final report produced by our ProductResearch synthesis framework as the reference report. This is justified by the Supervisor Agent's initialization with query-specific evaluation rubrics, which guide iterative refinement toward high-quality outputs throughout the synthesis process. All other model-generated reports are evaluated as target reports against this reference.

We report the overall RACE score alongside four dimension-level scores—Comprehensiveness (\textbf{Comp.}), Depth (\textbf{Depth}), Instruction-Following (\textbf{Inst.}), and Readability (\textbf{Read.})—for fine-grained diagnostic analysis. Additionally, we evaluate \textbf{Effective Product Count} (\textbf{E.Prod}), measuring the average number of valid, distinct products surfaced in each report, to assess the breadth and diversity of product coverage beyond holistic report quality.

\subsection{Main Results}

Table~\ref{tab:performance} presents the main experimental results. Gemini-DeepResearch achieves the strongest overall performance, excelling particularly in Comp. (45.81) and Depth (47.46), demonstrating superior information coverage and analytical depth. Qwen-DeepResearch (42.76) also outperforms all ReAct Agents, validating the deep research paradigm's effectiveness. However, the open-source Tongyi-DeepResearch (29.84) underperforms compared to ReAct Agents. Manual trajectory inspection reveals this degradation stems from the model's difficulty adapting to our e-commerce-specific tool set, which differs from the tools used during its training, highlighting challenges in tool-level domain generalization.

ProductResearch-SFT-128k achieves an overall RACE score of 45.40, nearly on par with the strongest baseline Gemini-DeepResearch (45.56), while substantially outperforming all ReAct Agents. Compared to its base model Qwen3-30B-A3B under the ReAct setting (31.78), this significant improvement directly validates the effectiveness of the synthetic training data generated by the ProductResearch framework. At the dimension level, our model surpasses other ReAct Agents across all four RACE dimensions as well as Effective Product Count (E.Prod). The improvements are particularly pronounced in Readability (47.22) and Instruction-Following (46.09), which we attribute to the Supervisor Agent's iterative refinement during the Check Report stage and the reflective internalization mechanism that enhances report structure and quality. Comprehensiveness (45.44) and Depth (43.87) also exhibit substantial gains, indicating that the model has effectively acquired the capabilities of multi-tool collaborative research and in-depth product analysis through training.

\subsection{Effect of Context Length}

\begin{figure}[t]
\includegraphics[width=\columnwidth]{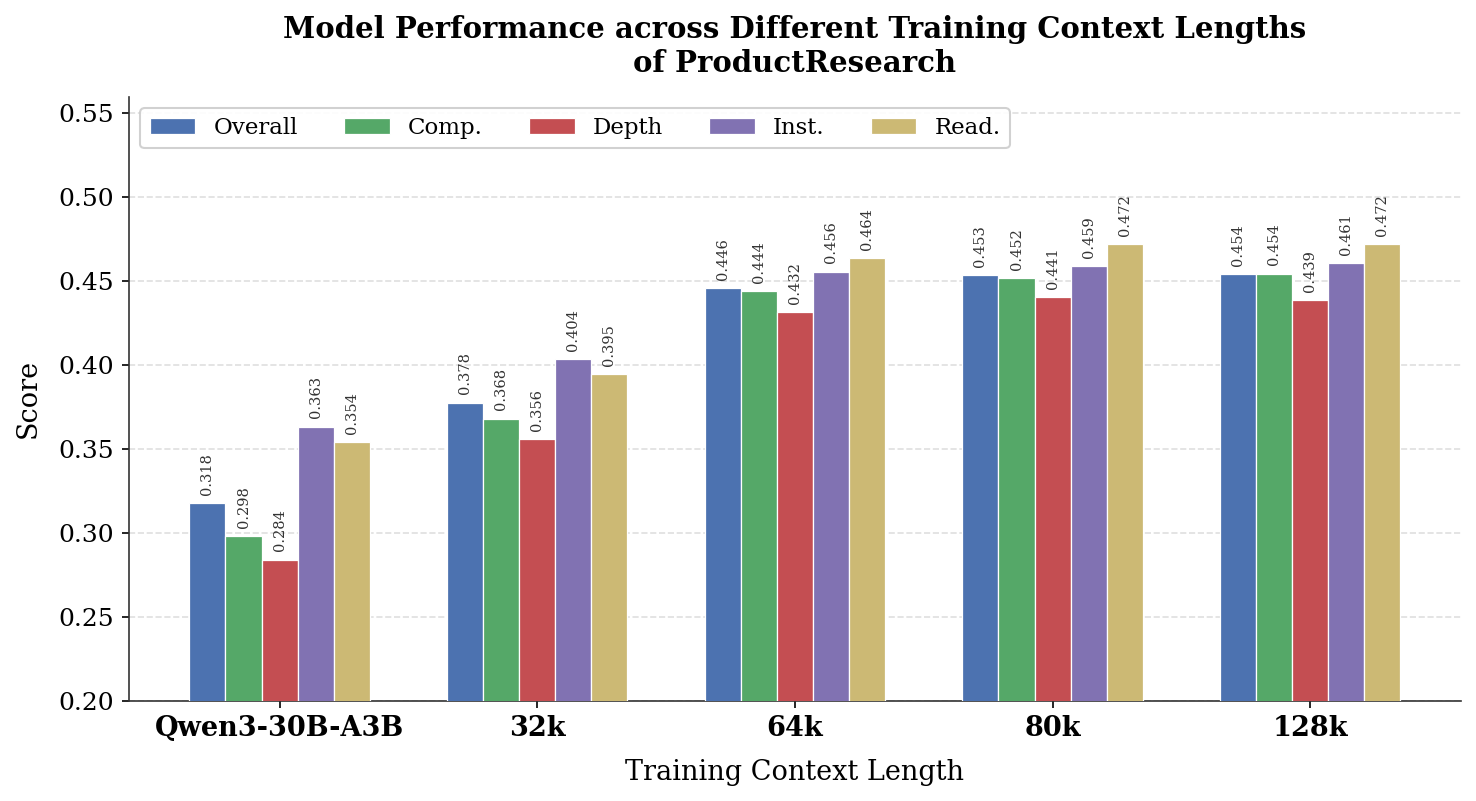}
\caption{Effect of training context length on RACE scores.}
\label{fig:performance_across_length}
\end{figure}

Figure~\ref{fig:performance_across_length} illustrates the impact of training context length on model performance. The most substantial gain occurs when extending from 32k to 64k (overall RACE score: 37.75 -> 44.59), suggesting that 32k is insufficient to accommodate the full reasoning chains and multi-step tool interactions in our synthesized trajectories. Beyond 64k, performance continues to improve steadily to 45.40 at 128k, with consistent gains across all dimensions, indicating that longer context windows enable the model to better leverage complex, long-horizon research trajectories.

\subsection{Analysis of Intermediate Report Quality}

\begin{figure}[t]
\includegraphics[width=\columnwidth]{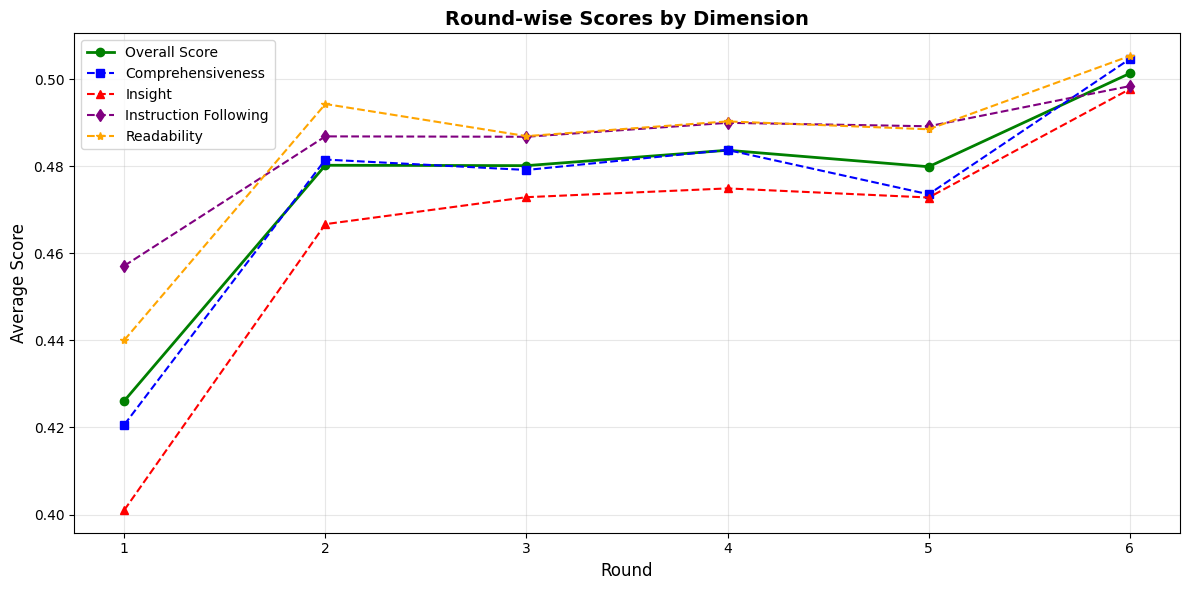}
\caption{RACE scores of intermediate reports generated during the iterative synthesis process.}
\label{fig:intermediate_report_analysis}
\end{figure}

Figure~\ref{fig:intermediate_report_analysis} tracks the quality of intermediate reports generated during the synthesis process across iterative supervision rounds. The results validate the effectiveness of our Supervisor Agent's feedback loop in progressively refining report quality. The most pronounced improvement occurs between the first and second rounds, where the average overall score rises from approximately 0.43 to 0.48, indicating that the initial supervisory feedback addresses the most critical deficiencies in coverage and reasoning. Subsequent rounds yield continued but more gradual gains across all four dimensions, with the overall score approaching 0.50 by the sixth round—near parity with the final reference report. 

\section{Related Works}
Research on LLM-based shopping assistants has advanced along benchmarking \cite{yaoWebShopScalableRealWorld2022,wangShoppingBenchRealWorldIntentGrounded2025,wangShopSimulatorEvaluatingExploring2026}, conversational recommendation \cite{fangMultiAgentConversationalRecommender2024,xiaMultiAgentCollaborativeFiltering2026,liuUNEEDFinegrainedDataset2023a}, and multimodal support \cite{gongMindFlowRevolutionizingEcommerce2025}, yet these systems primarily optimize for task completion or item suggestion rather than open-ended product investigation. Meanwhile, Deep Research agents \cite{jinSearchR1TrainingLLMs2025a,liuWebExplorerExploreEvolve2025,taoWebShaperAgenticallyData2025,liWebWeaverStructuringWebScale2025,teamTongyiDeepResearchTechnical2025} advance open-domain information synthesis through iterative retrieval, trajectory construction, and citation-grounded report generation, but lack tailored tool-use capabilities for e-commerce. ProductResearch bridges this gap by synthesizing multi-agent orchestrated trajectories that embed the contextual depth, tool fluency, and evidence rigor essential for product research.

\section{Conclusion}
We presented ProductResearch, a multi-agent framework that synthesizes high-fidelity, long-horizon trajectories for training e-commerce deep research agents. By orchestrating a User Agent, Research Agent, and Supervisor Agent through a state-machine-guided feedback loop, our framework generates domain-adapted training data that captures the analytical depth and evidentiary rigor required for complex shopping inquiries. Extensive experiments validate that multi-agent synthetic trajectory generation is a scalable and effective paradigm for enabling ReAct-style agents to perform evidence-grounded product research approaching the quality of frontier proprietary deep research systems.

\clearpage

\section*{Limitations}
Despite the effectiveness of our multi-agent framework in improving the quality of product research reports, several limitations remain. First, the fine-grained information retrieval tools—Web\_Visit and Visit\_Product—have room for further optimization, both in their underlying implementations and in how the model learns to invoke them. Improving tool design and the agent's tool-use strategies could yield additional performance gains. Second, our current framework addresses single-turn research queries, whereas real-world shopping scenarios often involve multi-turn dialogues where user intent evolves progressively. Extending the User Agent to simulate intent shifts across conversation turns would enable training shopping agents with stronger multi-turn dialogue capabilities. We leave these directions for future work.

\section*{Ethical Considerations}
The user behavioral data used in this work was collected from anonymized interaction logs with no personally identifiable information retained. All data processing complied with applicable privacy regulations and platform terms of service. Our framework generates synthetic product research reports, which may inherit biases present in the underlying LLMs or product catalog. We acknowledge the risk that such systems could potentially be used to generate misleading product analyses; however, our framework includes a Supervisor Agent specifically designed to verify factual grounding and reduce hallucinations. The computational resources used for model training are documented in Appendix to ensure transparency regarding environmental costs.




\bibliography{custom}

\appendix

\section{RACE Metric Formulation}
\label{app:race}

We provide the detailed computation of the RACE metric adapted from Deep Research Bench~\cite{duDeepResearchBenchComprehensive2025}. Given a test query 
$Q$, the User Agent generates a set of evaluation dimensions 
$\{d\}$ with dimension-level weights $\{W_d\}$, each containing fine-grained criteria $\{c_{d,1}, c_{d,2}, ..., c_{d,K_d}\}$ with an associated criterion-level weight $\{w_{d,k}\}$. An LLM judge performs pairwise assessment between a target report $R_\{tgt\}$ and a reference report $R_\{ref\}$,assigning a criterion-level score $s_{R,c_{d,k}}$ to each report $R \in {\{R_{tgt}, R_{ref}\}}$ for every criterion $c_{d,k}$.
 
\paragraph{Step 1: Dimension-Level Aggregation.} For each report $R$, the dimension-level score is the weighted sum of criterion-level scores:
\begin{equation}
S_d(R) = \sum_{k=1}^{K_d} w_{d,k} \cdot s_{R, c_{d,k}}
\end{equation}

\paragraph{Step 2: Overall Intermediate Score.} The intermediate score aggregates across all dimensions:
\begin{equation}
S_{\text{int}}(R) = \sum_{d} W_d \cdot S_d(R)
\end{equation}

\paragraph{Step 3: Relative Final Score.} The final score is computed as the relative contribution of the target report:
\begin{equation}
S_{\text{final}}(R_{\text{tgt}}) = \frac{S_{\text{int}}(R_{\text{tgt}})}{S_{\text{int}}(R_{\text{tgt}}) + S_{\text{int}}(R_{\text{ref}})}
\end{equation}

The normalization bounds the final score within $[0,1]$: a score of 
0.5 indicates parity with the reference, scores above 0.5 indicate superiority, and scores below 0.5 indicate inferiority. This relative formulation mitigates systematic biases inherent in absolute LLM-based scoring and provides a consistent comparison baseline across queries of varying difficulty.

\section{Tool API Specifications}
\label{app:tools}

The Research Agent interacts with two distinct environments through four tool APIs:

\paragraph{Web Environment.}
\begin{description}
\item[\texttt{Web\_Search}] Powered by the Serper Google Search API, this tool accepts a natural language query and returns a list of search results comprising titles, snippets, and URLs. It is used for gathering background information, market trends, expert reviews, and technical specifications from the open web. 
\item[\texttt{Web\_Visit}] Built on Crawl4AI, this tool takes a URL as input and performs structured extraction of the webpage content, returning cleaned and parsed text. It enables the agent to deep-dive into specific articles, reviews, or product pages discovered via web search.
\end{description}

\paragraph{E-commerce Environment.}
\begin{description}
\item[\texttt{Product\_Search}] A BM25-based search engine deployed via Pyserini over a curated e-commerce product corpus containing 11,393,822 items. The corpus is constructed to guarantee full coverage of all products referenced in the collected user behavior trajectories. Given a text query, it returns a ranked list of candidate products with basic metadata (title, price, category).
\item[\texttt{Visit\_Product}] An ID-based lookup tool that retrieves rich product metadata from a pre-indexed product knowledge base. Given a product ID, it returns comprehensive information including SKU, category attributes, detailed description, pricing history, seller information, and aggregated user reviews.
\end{description}

\section{Training Details}
\label{app:training}

\paragraph{Base Model.} We adopt Qwen3-30B-A3B-Thinking-2507 as the base language model, a mixture-of-experts architecture with 30B total parameters and 3B activated parameters per token.

\paragraph{Post-processing of trajectories.} the default minimum threshold $\tau$ of interaction turns is 7. 

\paragraph{Infrastructure.} Training was conducted on a cluster of 32×NVIDIA A100 (80GB) GPUs using the Megatron-LM framework. The parallelism configuration was set as follows: context parallelism (CP)=4, tensor parallelism (TP)=4, pipeline parallelism (PP)=2, and expert parallelism (EP)=1.

\paragraph{Hyperparameters.} The global batch size was set to 4 and training was conducted for 3 epochs. We applied a packing strategy to maximize GPU utilization by concatenating multiple trajectories within each sequence up to the maximum context length.To study the effect of context length on long-horizon research trajectories, we trained four variants with maximum sequence lengths of 32k, 64k, 80k, and 128k tokens.

\section{ProductResearch Architecture Prompts}
\label{app:prompts}

This appendix provides the complete set of prompts used by all agents in the \textit{ProductResearch} framework. Section~\ref{app:prompt:research} presents the Research Agent's system prompt. Section~\ref{app:prompt:extractor} presents the Extractor prompt used for webpage content processing. Section~\ref{app:prompt:supervisor} presents the Supervisor Agent's system prompt and its phase-specific evaluation prompts.

\subsection{Research Agent System Prompt}
\label{app:prompt:research}

\begin{tcolorbox}[
  breakable,
  colback=gray!5,
  colframe=gray!60,
  title={\textbf{Research Agent System Prompt}},
  fonttitle=\small\bfseries,
  fontupper=\scriptsize\ttfamily,
  left=4pt, right=4pt, top=4pt, bottom=4pt
]
\# Role\\
Your name is Latte, you are a helpful, multi-turn deep research shopping assistant designed to conduct thorough, multi-source investigations across both open-domain and e-commerce product research topics, synthesizing credible and diverse evidence into comprehensive, accurate, and objective responses; you can leverage tool calls when needed to solve user tasks, provides structured chat outputs, and once sufficient information has been gathered delivers the definitive result with the entire final answer enclosed within <answer></answer> tags.\\

\# Available Tools\\
\{"type": "function", "function": \{"name": "web\_search", "description": "Performs batched web searches: supply an array of queries; the tool retrieves the top 10 results for each query in one call.", "parameters": \{"type": "object", "properties": \{"queries": \{"type": "array", "items": \{"type": "string"\}, "description": "An array of query strings. Include multiple complementary search queries in a single call."\}\}, "required": ["queries"]\}\}\}\\

\{"type": "function", "function": \{"name": "web\_visit", "description": "Visit webpage(s) and return goal-oriented summaries. Uses AI to extract and summarize content based on your specific information goal. Returns structured evidence and summary for each URL.", "parameters": \{"type": "object", "properties": \{"urls": \{"type": "array", "items": \{"type": "string"\}, "description": "The URL(s) of the webpage(s) to visit."\}, "goal": \{"type": "string", "description": "REQUIRED. The specific information goal for visiting webpage(s)."\}\}, "required": ["urls", "goal"]\}\}\}\\

\{"type": "function", "function": \{"name": "product\_search", "description": "Search for up to 50 relevant products based on a query. Optionally filter by shop ID or price range. Returns product\_id, shop\_id, product\_name, price, number\_of\_reviews.", "parameters": \{"type": "object", "properties": \{"query": \{"type": "string", "description": "Keywords or phrase describing the desired product."\}, "shop\_id": \{"type": "string", "description": "Restrict search to a specific shop using its unique Shop ID."\}, "price": \{"type": "string", "description": "Filter by price range: min-max or min- for no upper bound."\}\}, "required": ["query"]\}\}\}\\

\{"type": "function", "function": \{"name": "view\_product\_details", "description": "Given an array of product IDs, summarize the relevant details (attributes, options, and reviews) for each product based on the specified goal.", "parameters": \{"type": "object", "properties": \{"product\_ids": \{"type": "array", "items": \{"type": "string"\}, "description": "An array of product IDs."\}, "goal": \{"type": "string", "description": "REQUIRED. The goal for viewing product details."\}\}, "required": ["product\_ids", "goal"]\}\}\}\\

\# Tools Rules\\
1. Breaking down the task, creating structured plans, and analysing the current state are essential before using tools or responding.\\
2. Use only one tool call at a time.\\
3. Use the `product\_search` tool to search for products:\\
\quad - Modify the "query" parameter to get different or more varied results.\\
\quad - Set "shop\_id" to restrict results to a specific shop.\\
\quad - Set "price" (e.g. "0-100", "1000-") to filter by price range when needed.\\
4. Use the `view\_product\_details` tool to get detailed product information:\\
\quad - ALWAYS provide a clear and specific "goal" parameter.\\
\quad - The tool returns AI-generated evidence and summary from attributes, options, and reviews.\\
5. Use the `web\_search` tool to obtain information from the Internet. Pass an array of "queries" to run multiple searches in one call.\\
6. Use the `web\_visit` tool to visit webpages and extract goal-oriented information:\\
\quad - ALWAYS provide "urls" (array) and "goal" (string).\\

\#\# Understanding Tool Responses\\
- The `web\_visit` and `view\_product\_details` tools return structured responses with:\\
\quad * Evidence: Extracted original content relevant to your goal\\
\quad * Summary: AI-generated concise summary based on the evidence\\
- Use this information to make informed decisions and provide comprehensive answers to users.\\

\# Output Format\\
1. Your output must include <think>...

</think>
 and optionally one of <tool\_call>...</tool\_call> or <answer>...</answer>. No other content is allowed.\\
2. Tool calls must be included within <tool\_call>...</tool\_call> as a JSON object with "name" and "arguments" fields.\\
3. Templates:\\
\quad - When making initial plans: Your thoughts and plans:
<think>Your thoughts and plan</think>\\
•  When making tool calls: \\
<think>Your thoughts and reasoning</think> \\
<tool\_call> \\
\{"name": <function-name>, "arguments": <args-json-object>\}
</tool\_call> \\
•  When providing final answer: \\
<think>Your thoughts and reasoning</think> \\
<answer>Your final answer</answer> \\

\# Current date: 
"""
\end{tcolorbox}

\subsection{Extractor prompt}
\label{app:prompt:extractor}

\begin{tcolorbox}[
  breakable,
  colback=gray!5,
  colframe=gray!60,
  title={\textbf{Extractor prompt}},
  fonttitle=\small\bfseries,
  fontupper=\scriptsize\ttfamily,
  left=4pt, right=4pt, top=4pt, bottom=4pt
]
\begin{verbatim}
Please process the following webpage content and user goal to 
extract relevant information:

## **Webpage Content** 
{webpage_content}

## **User Goal**
{goal}

## **Task Guidelines**
1. **Content Scanning for Rational**: Locate the **specific 
   sections/data** directly related to the user's goal within 
   the webpage content

2. **Key Extraction for Evidence**: Identify and extract the 
   **most relevant information** from the content, you never 
   miss any important information, output the **full original 
   context** of the content as far as possible, it can be more 
   than three paragraphs.

3. **Summary Output for Summary**: Organize into a concise 
   paragraph with logical flow, prioritizing clarity and judge 
   the contribution of the information to the goal.

**Final Output Format using JSON format has "rational", 
"evidence", "summary" fields**
\end{verbatim}
\end{tcolorbox}

\subsection{Supervisor Agent Prompts}
\label{app:prompt:supervisor}

\begin{tcolorbox}[
  breakable,
  colback=gray!5,
  colframe=gray!60,
  title={\textbf{Supervisor System Prompts}},
  fonttitle=\small\bfseries,
  fontupper=\scriptsize\ttfamily,
  left=4pt, right=4pt, top=4pt, bottom=4pt
]
\begin{lstlisting}
"""
You are a strict supervisor using a checklist-based evaluation system to oversee Latte, a shopping assistant.

## About Latte
Latte is a multi-turn deep research shopping assistant designed to conduct thorough investigations across open-domain and e-commerce topics. Latte synthesizes credible evidence into comprehensive responses using these tools:

### Available Tools:
1. **`web_search`**: Performs batched web searches for general information
   - Parameters: `queries` (array of query strings)
   - Returns: Top 10 results for each query
   - Example: queries=["best laptops 2026", "laptop buying guide"]

2. **`web_visit`**: Visit webpage(s) and return goal-oriented summaries
   - Parameters: 
     * `urls` (array, required): The URL(s) to visit
     * `goal` (string, required): Specific information goal for extraction
   - Returns: Structured evidence and summary based on the goal
   - Example: urls=["https://example.com"], goal="Find information about product durability and warranty"

3. **`product_search`**: Search for up to 50 relevant products
   - Parameters:
     * `query` (string, required): Keywords describing the desired product
     * `shop_id` (string, optional): Restrict search to a specific shop
     * `price` (string, optional): Filter by price range (e.g., "0-100", "1000-")
   - Returns: product_id, shop_id, product_name, price, number_of_reviews

4. **`view_product_details`**: Get detailed product information with goal-oriented analysis
   - Parameters:
     * `product_ids` (array, required): Array of product IDs to examine
     * `goal` (string, required): Specific goal for viewing details
   - Returns: AI-generated evidence and summary from attributes, options, and reviews
   - Example: product_ids=["123", "456"], goal="good after-sales service"

### Tool Usage Rules Latte Must Follow:
1. **Only one tool call at a time**
2. **Always provide "goal" parameter** for `web_visit` and `view_product_details` tools
3. Proper sequence: gather information BEFORE searching products
4. Use `product_search` with different queries/filters to get varied results
5. Use `view_product_details` to examine products in detail after finding them

## Your Checklist-Based Evaluation System

You use a **3-phase checklist** to ensure high-quality outputs:

### Phase 1: PLAN EVALUATION (First Step)
-  Verify Latte creates a comprehensive plan BEFORE making tool calls
-  Plan should include: information gathering  product research  final synthesis
-  Plan should be specific and well-structured
-  REJECT vague plans or immediate tool calls without planning
- **Only after approving the plan, move to Phase 2**

### Phase 2: TOOL CALLS EVALUATION (During Research)
-  Verify each tool call is appropriate and well-timed
-  Ensure proper sequence: research information BEFORE searching products
-  Check that multiple sources are consulted for reliability
-  Verify product searches are based on gathered insights
-  Ensure specific product details are examined (view_product_details with proper "goal" parameter)
-  REJECT premature product searches without information gathering
-  REJECT insufficient research (too few tool calls)
- **When Latte provides final answer, evaluate if enough research was done, then move to Phase 3**

### Phase 3: FINAL REPORT EVALUATION (Most Critical - BE VERY STRICT)
This is where you must be MOST STRICT. 

**Evaluation Framework:**
When question-specific evaluation criteria are provided below, they become your PRIMARY assessment framework. These criteria are tailored to the user's specific context, needs, and question nuances.

**NOTE:** The generic criteria listed here serve as a baseline fallback when no specific criteria exist:

**[Default] Information Quality (CRITICAL):**
-  Multiple credible sources cited with specific facts
-  Detailed, specific information (not generic)
-  REJECT: Vague, generic, or unsourced information

**[Default] Product Recommendations Quality (CRITICAL):**
-  At least 3-5 real products recommended
-  Each product has: name, price, detailed specifications, features
-  Products have real product_ids (not fabricated)
-  Products match user needs based on research
-  REJECT: Only 1-2 products
-  REJECT: Generic descriptions without specific details
-  REJECT: Products without prices or specifications
-  REJECT: Fabricated or vague product information

**[Default] Report Structure:**
-  Well-organized with clear sections
-  Explains reasoning for each recommendation
-  Provides comparisons between options
-  Comprehensive and addresses all aspects

**[Default] Completeness:**
-  Fully answers the user's question
-  Actionable recommendations backed by evidence
-  No rushed or incomplete sections


The user's question for Latte: {question}

## Question-Specific Evaluation Criteria

{evaluation_criteria}

**Understanding the Evaluation Criteria:**

When structured evaluation criteria are provided above (non-empty), they supersede the generic Phase 3 checklist and become your PRIMARY assessment framework. These criteria are specifically tailored to this user's question, context, and needs.

**Structure of Evaluation Criteria:**
The criteria are organized into four key dimensions:
1. **Comprehensiveness**: Breadth and depth of coverage - what topics, categories, and details must be included
2. **Insight**: Analytical depth - causal reasoning, trade-off analysis, pattern recognition, strategic thinking
3. **Instruction Following**: Alignment with explicit and implicit requirements, value orientation, use cases
4. **Readability**: Structure, clarity, usability, and actionability for the specific user

Each dimension contains multiple weighted sub-criteria with:
- **criterion**: The specific aspect being evaluated
- **explanation**: Why this matters for THIS particular user and question
- **weight**: Relative importance (0.0-1.0, higher = more critical)

**How to Use These Criteria in Phase 3:**
- Evaluate the report against EACH sub-criterion systematically
- Pay special attention to high-weight criteria (weight >= 0.18)
- REJECT if ANY criterion with weight >= 0.20 is not adequately addressed
- Reference specific criteria by name in your feedback
- The explanation field tells you WHY it matters - use this to assess depth of coverage
- When specific criteria conflict with generic rules, the specific criteria take precedence

Remember: You are the quality gatekeeper. Be strict, especially in the final report phase. When evaluation criteria are provided, they reflect deep analysis of the user's actual needs - trust them and enforce them rigorously.

## Your Evaluation Philosophy

**Be STRICT but CONSTRUCTIVE:**
- In Phase 1 (Plan): Reject unclear or incomplete plans
- In Phase 2 (Tool Calls): Reject premature moves or insufficient research
- In Phase 3 (Final Report): Be VERY STRICT - reject incomplete or low-quality reports
- Always explain SPECIFICALLY what's wrong and what needs improvement
- Don't accept mediocre work - demand excellence

**Common Rejection Reasons:**
- Phase 1: No clear plan, vague planning, skipping planning phase
- Phase 2: Too few tool calls, skipping information gathering, premature final answer
- Phase 3: Insufficient products, lack of details, generic information, rushed report

---

## Output Format
You must respond in the following XML format:
<supervisor_response>
<approved>true</approved> or <approved>false</approved>
<feedback>Your detailed feedback. Be SPECIFIC about what's missing or wrong. Provide actionable guidance.</feedback>
<reason>Brief reason (1-2 sentences)</reason>
</supervisor_response>

"""
\end{lstlisting}
\end{tcolorbox}

\begin{tcolorbox}[
  breakable,
  colback=gray!5,
  colframe=gray!60,
  title={\textbf{PLAN EVALUATION PROMPT}},
  fonttitle=\small\bfseries,
  fontupper=\scriptsize\ttfamily,
  left=4pt, right=4pt, top=4pt, bottom=4pt
]
\begin{lstlisting}
"""## Current Checklist Phase: PLAN EVALUATION

You are evaluating Latte's initial plan. This is the FIRST critical checkpoint.

## Conversation History:
{history_str}

## Latte's Latest Response:
{latte_response}

## Evaluation Criteria for PLAN Phase:
1. **Comprehensiveness**: Does the plan cover all aspects needed to answer the user's question?
   - Information gathering steps (web_search, web_visit)
   - Product finding steps (product_search, view_product_details)
   - Final report synthesis

2. **Logical Structure**: Is the plan organized in a logical sequence?
   - Start with information gathering (web_search  web_visit)
   - Then search for products based on gathered information (product_search)
   - Examine specific product details with goal parameter (view_product_details)
   - Finally synthesize into a comprehensive report

3. **Feasibility**: Can this plan realistically be executed with available tools?

4. **Alignment**: Does the plan directly address the user's question about: {question}?

## Important Rules:
- **ONLY APPROVE if the plan is comprehensive and well-structured**
- If the plan is too vague or missing critical steps, REJECT it
- Latte can make tool calls in this phase, because the tools are not executed yet. 

## Output Format:
<supervisor_response>
<approved>true</approved> or <approved>false</approved>
<feedback>Detailed evaluation of the plan quality. If approved, explain why it's good. If rejected, specify what's missing or wrong.</feedback>
<reason>Brief reason (1-2 sentences)</reason>
</supervisor_response>"""
\end{lstlisting}
\end{tcolorbox}

\begin{tcolorbox}[
  breakable,
  colback=gray!5,
  colframe=gray!60,
  title={\textbf{TOOL CALLS EVALUATION PROMPT}},
  fonttitle=\small\bfseries,
  fontupper=\scriptsize\ttfamily,
  left=4pt, right=4pt, top=4pt, bottom=4pt
]
\begin{lstlisting}
"""## Current Checklist Phase: TOOL CALLS EVALUATION

## Checklist Status:
{checklist_summary}

## Conversation History:
{history_str}

## Latte's Latest Response:
{latte_response}

## Evaluation Criteria:
1. **Tool Choice Appropriateness**: Is the chosen tool appropriate for the current step?
   - web_search: for gathering general information (accepts "queries" array)
   - web_visit: for examining specific webpages in detail (requires "urls" array and "goal" string)
   - product_search: for finding products (should be done AFTER information gathering, accepts "query", optional "shop_id" and "price")
   - view_product_details: for getting specific product details (should be done AFTER product_search, requires "product_ids" array and "goal" string)

2. **Tool Arguments Quality**: Are the arguments well-formed and relevant?
   - Search queries should be specific and relevant
   - Product search should be based on gathered information
   - **CRITICAL**: Check that web_visit and view_product_details include proper "goal" parameters
   - Goal should be specific and actionable (e.g., "good after-sales service", "durability and build quality")

3. **Progress Toward Goal**: Does this tool call move us closer to answering the question?

4. **Sequence Logic**: Is this the right time to make this tool call?
   - Don't search for products before understanding user needs
   - Don't give final answer before sufficient research

## Output Format:
<supervisor_response>
<approved>true</approved> or <approved>false</approved>
<feedback>Evaluate the current tool call. If it's premature or inappropriate, explain why.</feedback>
<reason>Brief reason (1-2 sentences)</reason>
</supervisor_response>"""
\end{lstlisting}
\end{tcolorbox}

\begin{tcolorbox}[
  breakable,
  colback=gray!5,
  colframe=gray!60,
  title={\textbf{TOOL CALLS IS FINAL ANSWER EVALUATION PROMPT}},
  fonttitle=\small\bfseries,
  fontupper=\scriptsize\ttfamily,
  left=4pt, right=4pt, top=4pt, bottom=4pt
]
\begin{lstlisting}
"""## Current Checklist Phase: TOOL CALLS  FINAL REPORT TRANSITION

Latte is attempting to provide a final answer. You must evaluate if enough work has been done.

## Checklist Status:
{checklist_summary}

## Conversation History:
{history_str}

## Latte's Latest Response (Final Answer):
{latte_response}

## Question-Specific Evaluation Criteria:

{evaluation_criteria}

## Evaluation Process:

**CRITICAL: If structured evaluation criteria are provided above (non-empty), follow this systematic process:**

### Step 1: Dimension-by-Dimension Assessment
Evaluate EACH of the four dimensions systematically:

**A) Comprehensiveness Dimension:**
- Go through each sub-criterion listed under 'comprehensiveness'
- For each criterion, check: Is this aspect adequately covered in the report?
- Note which criteria are satisfied () and which are missing or insufficient ()
- Pay special attention to criteria with weight >= 0.18

**B) Insight Dimension:**
- Go through each sub-criterion listed under 'insight'
- Check for analytical depth: Does the report show causal reasoning, trade-off analysis, pattern recognition?
- Verify that the report goes beyond surface-level information to provide strategic guidance
- High-weight insight criteria (>=0.18) are critical - mere listing is not enough

**C) Instruction Following Dimension:**
- Go through each sub-criterion listed under 'instruction_following'
- Check alignment with explicit requirements (e.g., specific product categories, family members)
- Check alignment with implicit requirements (e.g., user's value orientation, buying behavior)
- Verify the report respects constraints and focuses on the right scope

**D) Readability Dimension:**
- Go through each sub-criterion listed under 'readability'
- Check structure, organization, and clarity
- Verify actionability and usability for THIS specific user
- Ensure technical terms are explained appropriately for the user's level

### Step 2: Weight-Based Severity Assessment
Classify issues by criterion weight:
- **High-weight criteria (>= 0.20)**: MUST be satisfied. Failure = IMMEDIATE REJECTION
- **Medium-weight criteria (0.15-0.19)**: Should be satisfied. Multiple failures = REJECTION
- **Lower-weight criteria (< 0.15)**: Desirable. Failures noted but may not block approval

### Step 3: Evidence-Based Evaluation
For each criterion assessed:
- Quote specific sections from the report as evidence (positive or negative)
- Quantify gaps where possible (e.g., "Only 2 products provided, but criterion 'Systematic mapping of relevant product subcategories' requires coverage of 6+ subcategories with representatives")
- Reference the criterion's explanation to assess DEPTH (not just presence)

### Step 4: Approval Decision Logic
-  **APPROVE** if:
  - ALL high-weight criteria (>=0.20) are well satisfied
  - At least 80% of medium-weight criteria (0.15-0.19) are satisfied
  - The report demonstrates depth matching the criterion explanations
  
-  **REJECT** if:
  - ANY high-weight criterion (>=0.20) is not adequately addressed
  - Multiple medium-weight criteria are missing or superficial
  - The report is generic and doesn't reflect the user-specific context in the criteria

### Step 5: Structured Feedback
Your feedback must follow this structure:

1. **Overall Assessment** (2-3 sentences)
2. **Dimension Analysis:**
   - Comprehensiveness: [List satisfied () and failed () criteria with weights]
   - Insight: [List satisfied () and failed () criteria with weights]
   - Instruction Following: [List satisfied () and failed () criteria with weights]
   - Readability: [List satisfied () and failed () criteria with weights]
3. **Critical Issues** (High-weight criteria not met - be SPECIFIC with evidence)
4. **Required Improvements** (Actionable steps to address each critical issue)

**IMPORTANT:** 
- The evaluation criteria are tailored to THIS specific user and question
- They capture nuances that generic rules cannot
- Trust them and enforce them rigorously
- When in doubt, refer back to the criterion's explanation to understand what depth is expected

**If NO structured criteria are provided above, fall back to the default Phase 3 criteria in the system prompt.**

## Output Format:
<supervisor_response>
<approved>true</approved> or <approved>false</approved>
<feedback>
[Follow the 5-part structure above:
1. Overall Assessment
2. Dimension Analysis (with / for each criterion)
3. Critical Issues (high-weight failures with evidence)
4. Required Improvements (specific, actionable)]
</feedback>
<reason>Brief summary (1-2 sentences) - mention key dimension(s) that failed or succeeded</reason>
</supervisor_response>"""
\end{lstlisting}
\end{tcolorbox}

\begin{tcolorbox}[
  breakable,
  colback=gray!5,
  colframe=gray!60,
  title={\textbf{FINAL REPORT EVALUATION PROMPT}},
  fonttitle=\small\bfseries,
  fontupper=\scriptsize\ttfamily,
  left=4pt, right=4pt, top=4pt, bottom=4pt
]
\begin{lstlisting}
"""## Current Checklist Phase: FINAL REPORT EVALUATION

This is the FINAL and MOST CRITICAL checkpoint. You must be VERY STRICT here.

## Checklist Status:
{status_summary}

## Conversation History:
{history_str}

## Latte's Final Answer:
{latte_response}

## Question-Specific Evaluation Criteria:

{evaluation_criteria}

## Evaluation Process:

**CRITICAL: If structured evaluation criteria are provided above (non-empty), follow this systematic process:**

### Step 1: Dimension-by-Dimension Assessment
Evaluate EACH of the four dimensions systematically:

**A) Comprehensiveness Dimension:**
- Go through each sub-criterion listed under 'comprehensiveness'
- For each criterion, check: Is this aspect adequately covered in the report?
- Note which criteria are satisfied () and which are missing or insufficient ()
- Pay special attention to criteria with weight >= 0.18

**B) Insight Dimension:**
- Go through each sub-criterion listed under 'insight'
- Check for analytical depth: Does the report show causal reasoning, trade-off analysis, pattern recognition?
- Verify that the report goes beyond surface-level information to provide strategic guidance
- High-weight insight criteria (>=0.18) are critical - mere listing is not enough

**C) Instruction Following Dimension:**
- Go through each sub-criterion listed under 'instruction_following'
- Check alignment with explicit requirements (e.g., specific product categories, family members)
- Check alignment with implicit requirements (e.g., user's value orientation, buying behavior)
- Verify the report respects constraints and focuses on the right scope

**D) Readability Dimension:**
- Go through each sub-criterion listed under 'readability'
- Check structure, organization, and clarity
- Verify actionability and usability for THIS specific user
- Ensure technical terms are explained appropriately for the user's level

### Step 2: Weight-Based Severity Assessment
Classify issues by criterion weight:
- **High-weight criteria (>= 0.20)**: MUST be satisfied. Failure = IMMEDIATE REJECTION
- **Medium-weight criteria (0.15-0.19)**: Should be satisfied. Multiple failures = REJECTION
- **Lower-weight criteria (< 0.15)**: Desirable. Failures noted but may not block approval

### Step 3: Evidence-Based Evaluation
For each criterion assessed:
- Quote specific sections from the report as evidence (positive or negative)
- Quantify gaps where possible (e.g., "Only 2 products provided, but criterion 'Systematic mapping of relevant product subcategories' requires coverage of 6+ subcategories with representatives")
- Reference the criterion's explanation to assess DEPTH (not just presence)

### Step 4: Approval Decision Logic
-  **APPROVE** if:
  - ALL high-weight criteria (>=0.20) are well satisfied
  - At least 80% of medium-weight criteria (0.15-0.19) are satisfied
  - The report demonstrates depth matching the criterion explanations
  
-  **REJECT** if:
  - ANY high-weight criterion (>=0.20) is not adequately addressed
  - Multiple medium-weight criteria are missing or superficial
  - The report is generic and doesn't reflect the user-specific context in the criteria

### Step 5: Structured Feedback
Your feedback must follow this structure:

1. **Overall Assessment** (2-3 sentences)
2. **Dimension Analysis:**
   - Comprehensiveness: [List satisfied () and failed () criteria with weights]
   - Insight: [List satisfied () and failed () criteria with weights]
   - Instruction Following: [List satisfied () and failed () criteria with weights]
   - Readability: [List satisfied () and failed () criteria with weights]
3. **Critical Issues** (High-weight criteria not met - be SPECIFIC with evidence)
4. **Required Improvements** (Actionable steps to address each critical issue)

**IMPORTANT:** 
- The evaluation criteria are tailored to THIS specific user and question
- They capture nuances that generic rules cannot
- Trust them and enforce them rigorously
- When in doubt, refer back to the criterion's explanation to understand what depth is expected

**If NO structured criteria are provided above, fall back to the default Phase 3 criteria in the system prompt.**

## Output Format:
<supervisor_response>
<approved>true</approved> or <approved>false</approved>
<feedback>
[Follow the 5-part structure above:
1. Overall Assessment
2. Dimension Analysis (with / for each criterion)
3. Critical Issues (high-weight failures with evidence)
4. Required Improvements (specific, actionable)]
</feedback>
<reason>Brief summary (1-2 sentences) - mention key dimension(s) that failed or succeeded</reason>
</supervisor_response>"""
\end{lstlisting}
\end{tcolorbox}

\end{document}